\documentclass{article}
\usepackage{spconf,amsmath,graphicx,amsfonts}

\title{Shuffled Autoregression For Motion Interpolation}
\name{Shuo Huang$^{1}$, Jia Jia$^{1}$$^{3}$, Zongxin Yang$^{2}$, Wei Wang, Haozhe Wu$^{1}$, Yi Yang$^{2}$, Junliang Xing$^{1}$$^\ast$\thanks{*Corresponding author}}
\address{$^{1}$Department of Computer Science and Technology, Tsinghua University, Beijing 100084, China \\
$^{2}$CCAI, College of Computer Science and Technology, Zhejiang University\\
$^{3}$Beijing National Research Center for Information Science and Technology
}
\begin{document}
\ninept
\maketitle
\begin{abstract}
%兴军亮：目前摘要研究动机铺垫部分有些啰嗦了，最好通过一到两句话就直接引入到要做的内容；没有突出自己方法的创新性，也没有比较清楚地让读者理解你的做法和步骤；Shuffled AutoRegressio没有解释清楚，名字叫AutoRegression，什么又说可以同时进行AutoRegression和Non-AutoRegreesion？改成HybridRegression？
This work aims to provide a deep-learning solution for the motion interpolation task. Previous studies solve it with geometric weight functions. Some other works propose neural networks for different problem settings with consecutive pose sequences as input. However, motion interpolation is a more complex problem that takes isolated poses (e.g., only one start pose and one end pose) as input. When applied to motion interpolation, these deep learning methods have limited performance since they do not leverage the flexible dependencies between interpolation frames as the original geometric formulas do. To realize this interpolation characteristic, we propose a novel framework, referred to as \emph{Shuffled AutoRegression}, which expands the autoregression to generate in arbitrary (shuffled) order and models any inter-frame dependencies as a directed acyclic graph. We further propose an approach to constructing a particular kind of dependency graph, with three stages assembled into an end-to-end spatial-temporal motion Transformer. Experimental results on one of the current largest datasets show that our model generates vivid and coherent motions from only one start frame to one end frame and outperforms competing methods by a large margin. The proposed model is also extensible to multiple keyframes' motion interpolation tasks and other areas' interpolation.

\end{abstract}

\section{Introduction}
\label{sec:intro}

When creating character animations, animators use human motion interpolation methods to fill blank frames between hand-made keyframes. 
The general methods provided by animation software, such as spline curves \cite{ferguson1964}, fail to depict vivid human motions.
To achieve better performance, previous researchers attempt to improve the interpolation kernels of the weight functions from two perspectives. 
They either use general interpolation functions in higher order space, like QLERP \cite{Kavan2005}, SpFus \cite{patron2015spline}, or establish a human joint model and use statistical methods to study it \cite{Mukai2005}.
However, in many cases, these mathematical methods still need manually editing of curve parameters or more keyframes to obtain better visual quality.
% However, these methods lack the expressiveness of neural networks, and they requires manual editing of curve parameters or more keyframes for better visual quality, resulting in a lot of workforce consumption.
% However, current methods provided by animation software, such as spline curves \cite{ferguson1964}, fail to depict vivid human motions. So it requires manual editing of curve parameters or more keyframes for better visual quality, resulting in a lot of workforce consumption. 
% The goal of our research is to release human labor by proposing a specific interpolation method for human motion that is data-driven.

\begin{figure}[t]
\centering
    \includegraphics[scale=0.24]{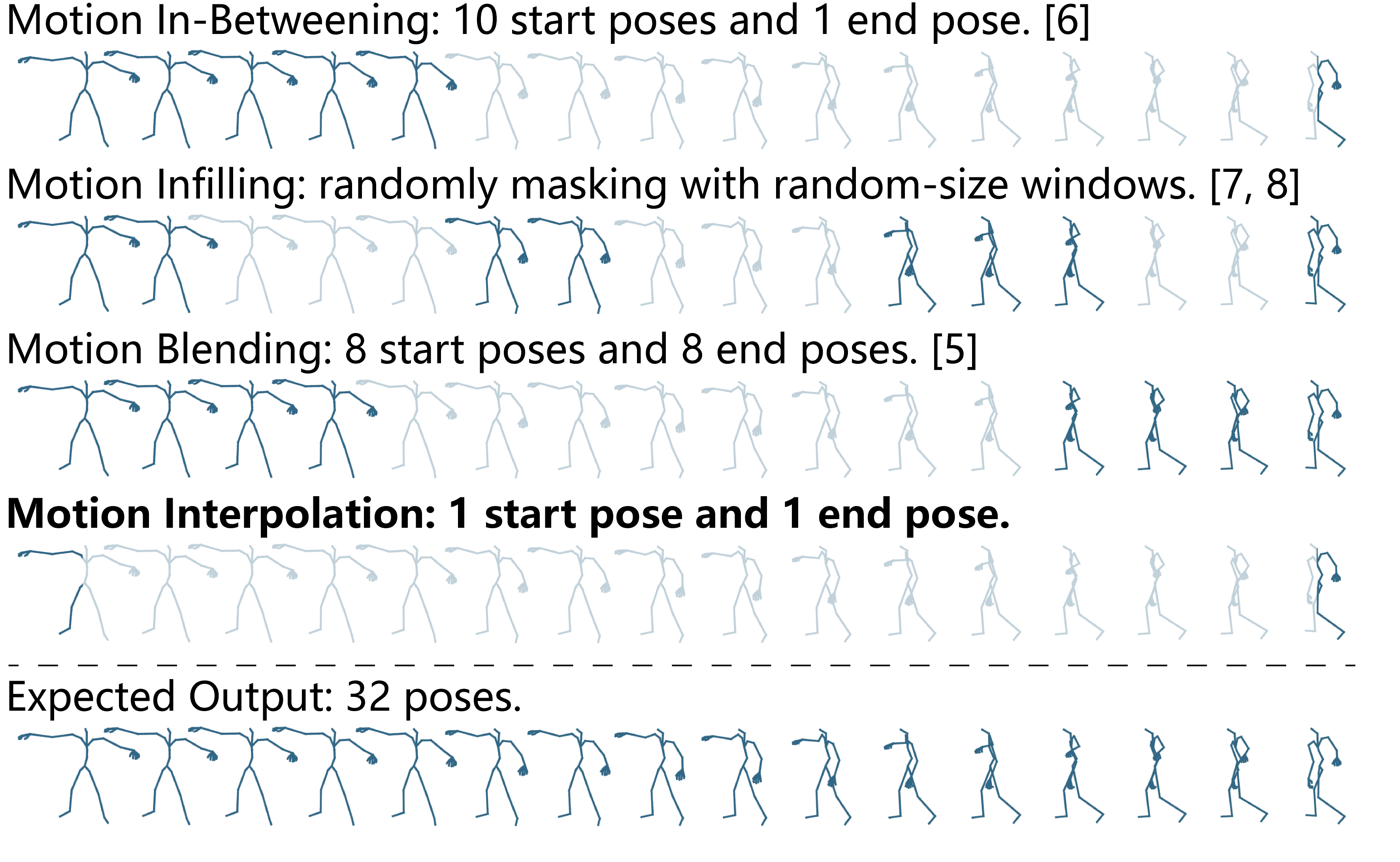}
    \caption{The differences between input proportions (the unmasked proportions) of motion completion scenarios (motion in-betweening, infilling, blending, interpolation). Please note that the example motion consists of 32 poses, and one pose drawn in this figure stands for two poses in the motion sequence.}
    \label{fig:completion}
\end{figure}

% 之前的工作已经表明motion interpolation是rational。存在两个问题: (1 ) 误差累积 (2) 过平滑,，一部分工作是按某种特定顺序做回归，另一部分工作是并行生成，例如xxx方法(cite)。
Recently several works \cite{duan2022, harvey2020, holden2016, kaufmann2020} have contributed to similar tasks. Including motion interpolation, these tasks are collectively referred to as motion completion. The difference between their inputs is shown in Fig.~\ref{fig:completion}.
For the in-betweening task, \cite{harvey2020} propose time-to-arrival embedding and scheduled target noise, forming a robust RNN-based autoregressive model.
For the infilling task, \cite{holden2016, kaufmann2020} view human motion infilling as spatial-temporal image inpainting and apply autoencoders to reconstruct motions.
For both tasks and the blending task, \cite{duan2022} use a Transformer \cite{Vaswani2017, devlin2019} encoder to modify the linear interpolation, which is a more versatile method. 

Though previous deep learning methods have achieved
promising results on these similar tasks, there are still some challenges to be solved. 
On the one hand, the above methods show poor performance when applied to motion interpolation, which is a more complex problem with the smallest input size.
Due to the substantial reduction of input information, autoregressive models suffer from severe error accumulation, which hinders them from finishing the transition to the end pose. In contrast, non-autoregressive models tend to predict over-smoothed motions.
On the other hand, these methods don't leverage some essence of the interpolation task. The weight function method \cite{ferguson1964,Kavan2005,patron2015spline,Mukai2005} shows that it is reasonable to generate an interpolation frame with any frame as the condition. Given the linear interpolation, the impact of each keyframe on the resulting frame depends only on their temporal distance. So it is acceptable that the keyframe is in the past or the future, away from or near the resulting frame. This indicates that the frames have flexible dependencies in an interpolation task.

%We think that the essence of problems is that previous methods cannot fully adapt to the special relationship between interpolation frames. When generating a specific frame in interpolation tasks, future and past information are of the same importance, depending on the temporal distance. The previous autoregressive methods focus more on past information as if they are solving a prediction task.

\begin{figure*}[t]
\centering
    \includegraphics[scale=0.45]{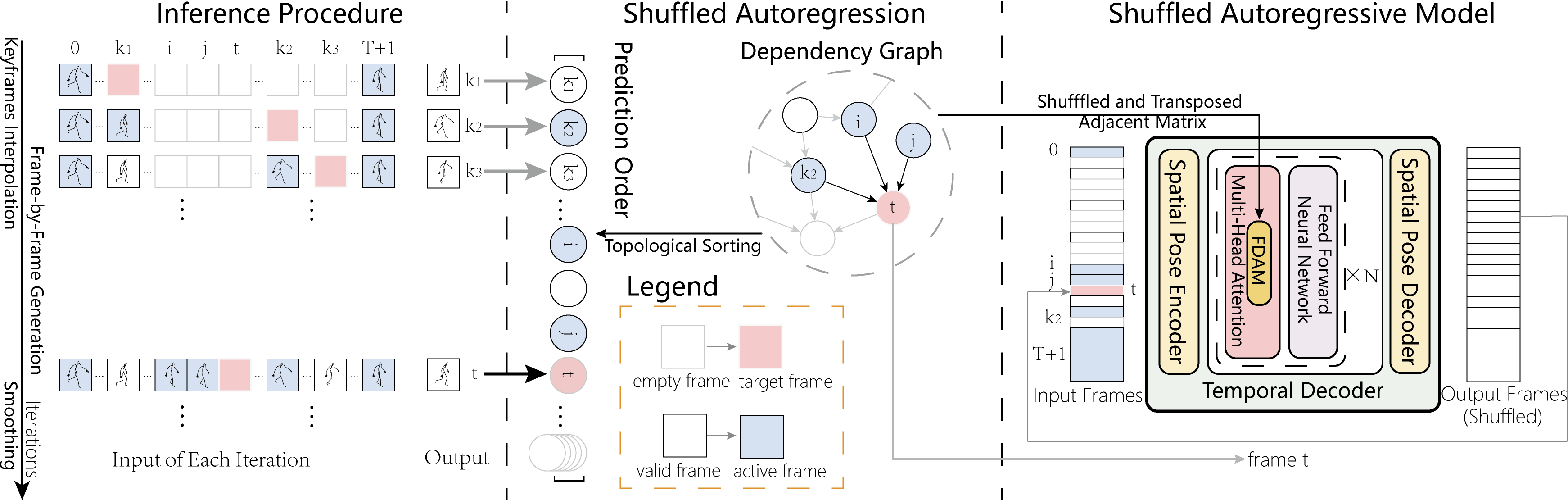}
    \caption{From left to right are the inference procedure of the SAR, the SAR framework, and the SAR model. The Legend is in the lower middle. An empty frame means that we haven't generated this frame yet. When the current iteration is generating it, it becomes a target frame. After its generation, it becomes valid. If the ongoing iteration depends on a valid frame, then the frame becomes active. 
    Firstly, we design a dependency graph whose topology can be explained as the pipeline of the inference procedure. ($k_n$ in the first stage means the n-th keyframe.) Secondly, we construct the SAR model, deriving the FDAM from the graph. Thirdly, the model works in iterations of the inference procedure. The middle and right parts of the figure shows the iteration when predicting frame $t$ conditioned on frame $0$, $k_2$, $i$, $j$, $T+1$. We can find the target frame for every iteration by querying the dependency graph. Then we put it back at the input for the next iteration.}
    \label{fig:overview}
\end{figure*}
To address the problems above, we propose a Shuffled AutoRegression (SAR) method. It is an extension of the original autoregressive (AR) framework. The AR framework generates future frames from past frames in chronological order. In contrast, as shown in Fig.~\ref{fig:overview}, the SAR framework generates frames in a custom order, and the dependencies between frames are also freely selected, which enables it to catch the flexible inter-frame dependencies. In fact, the SAR framework can express any dependency as long as these dependencies form a directed acyclic graph.
For the first problem, SAR alleviates the error accumulation of autoregressive methods and takes advantage of parallel generation. The error accumulates along the dependency graph, so we can manipulate its topology to constrain it.

In particular, we design a DAG topology representing a motion interpolation pipeline. The proposed pipeline contains three stages: 1) keyframe interpolation, 2) frame-by-frame generation, and 3) motion smoothing. The three stages follow under a divide-and-conquer strategy to constraint error accumulation. We incorporate the three stages in an end-to-end Transformer \cite{Vaswani2017} architecture, the attention mechanism of which describes the direct relationships between any two frames. We prove that the architecture is fit to carry out the SAR by controlling the attention mask. Specifically, we propose a Flexible Dependency Attention Mask (FDAM) module, which enables a Transformer Decoder (GPT-like) to carry out SAR.
% As shown in Fig.~\ref{fig:overview}, the shuffled order of SAR corresponds to a Directed Acyclic Graph (DAG) of prediction dependencies and the order is the topological sorting array of the DAG. In this way, the error accumulates along the DAG, so we can manipulate the topology to constrain it.
% to address the aforementioned problems, 我们提出了一种xxx方法，which同时借鉴了并行生成和回归的优点。
% 我们的核心思路是xxx。(分而治之，先插更多帧，再逐帧生成)

% More specifically, we design a topology that represents a pipeline for motion interpolation. The pipeline contains three stages: (1) keyframe interpolation, (2) frame-by-frame generation, and (3) motion smoothing. We incorporate the 3 stages in an end-to-end Motion Transformer.
% The three stages follow under a divide-and-conquer strategy to constraint error accumulation. Moreover, these three stages are a general solution to interpolation, which can be applied to interpolations of other areas.
% 再说由于transformer的attentionxxx之类的特点，在做shuffled autogression更有优势。

To evaluate the effectiveness of our model, we evaluate our model for motion interpolation between two frames on the massive AMASS dataset \cite{Mahmood2019}. To enforce our model's generality, we embrace motions of various amplitudes and design different sliding windows for different framerates. Experiments show the effectiveness of our proposed model compared with state-of-the-art deep-learning methods. Although the interpolation between two frames is discussed later in this paper, the flexibility of the SAR makes it easy to extend the model to multiple frames' interpolation.

The contributions of the paper can be summarized as follows:
\begin{itemize}
    \item
    We propose a deep learning method to solve motion interpolation, a more complex setting than other works. And it outperforms state-of-the-art methods for motion completion.
    \item 
    We propose to address the motion interpolation problem with Shuffled AutoRegression architecture and an end-to-end model realizing the SAR. The approach can be extended to interpolation problems in other fields.
    \item
    We propose a Flexible Dependency Attention Mask (FDAM) module, which enables a Transformer to carry out the SAR.
\end{itemize}  

\section{Problem Formulation}
Our human body's skeletal structure consists of rigid bone segments linked by $J$ joints, where each joint consists of a relative rotation angle of 3 rotational Degrees of Freedom (DoF). We shall represent a pose by a tensor $\mathbf{p} \in \mathbb{R}^{J \times 3}$. Given the pose at the start moment $\mathbf{p}^{(0)}$ and the end moment $\mathbf{p}^{(T+1)}$, motion interpolation is to generate a sequence of poses $\mathbf{P}=[\mathbf{p}^{(1)}, \mathbf{p}^{(2)}, \dots ,\mathbf{p}^{(T)}] \in \mathbb{R} ^ {T \times J \times 3}$. The generated motion should accord with human exercise habits and directly connect the given frames. As this problem is non-deterministic, in practice, the results are often optimized to fit into the existing motion data, indicated by the distance of each generated pose from the ground truth. As the bone length of our body model is simplified and inherently encoded, here we optimize the L2 distance of joint angles between generation and the ground truth to a minimum.

\section{Proposed Method}
\label{sec:method}

Aiming to reduce the error accumulation of autoregressive models, we propose Shuffled AutoRegression (SAR), whose dependencies form a Directed Acyclic Graph (DAG). (If frame $i$ conditions on the prediction of frame $j$, then there exists \textbf{dependency} from frame $i$ to frame $j$.) The error accumulates along the edges of the dependency graph.
Based on this observation, we design a specific DAG for motion interpolation in which error accumulates much slower than the original order, as shown in Fig.~\ref{fig:overview}. The designed order consists of three stages, which form a pipeline for the task. Note that the SAR can handle any kind of DAG. Here we only state a particular kind of DAG that converges quickly and works well in the attempt. And we compare it with a DAG representing binary search in subsection~\ref{subsec:ablation}.

To carry out the SAR, we leverage the Transformer \cite{Vaswani2017} decoder used in GPT-2 \cite{Alec2019} as the backbone of the regressor. Here, we design a Flexible Dependency Attention Mask (FDAM) module, which enables the architecture to fit SAR. We further prove that FDAM is a shuffled and transposed adjacent matrix of the dependency graph.

\subsection{Shuffled AutoRegression} \label{subsection:SAR}
\subsubsection{Prediction Order and Dependency Graph}
Before introducing the concept of SAR, we first elaborate on why the error accumulates so fast in the original autoregressive order. To generate a sequence $\mathbf{X}=[\mathbf{x}_{0}, \mathbf{x}_{1},\mathbf{x}_{2}, \ldots, \mathbf{x}_{T-1}]$ ($\mathbf{x}_{\mathrm{inp}}$ as default input), the autoregressive prediction order is a sequential order:
\begin{equation}
    \mathbf{O}_{AR} = [0, 1, 2,...,T-1] ,\label{1}
\end{equation}
\begin{equation}
    \mathbf{x}_{i} = \mathrm{f}( \mathbf{x}_{\mathrm{inp}}, \mathbf{x}_{0}, \mathbf{x}_{1}, \mathbf{x}_{2}, \ldots, \mathbf{x}_{i-1}; \Theta), \label{2}
\end{equation}
where $\mathrm{f}$ is the function of an autoregressive model with parameters $\Theta$. However, in motion interpolation, most of the input frames to the model are biased predictions, while only two of them are the ground truth. This leads to an unacceptable error accumulation speed.

We propose Shuffled AutoRegression (SAR) to alleviate the problem. We only choose those elements that are highly related and have fewer errors as references for the following generation to reduce the proportion of deviation. And to make those highly correlated elements happen to be the ones with the minor error, SAR breaks down the left-to-right order along with the half-full dependencies. Our formulas are shown below:
\begin{equation}
    \mathbf{O}_{SAR} = [ t_{0}, t_{1}, t_{2}, \ldots, t_{T-1}], \label{4}
\end{equation}
\begin{equation}
    \mathbf{x}_{t_{i}} = \mathrm{f}(\mathbf{x}_{\mathrm{inp}}, \mathrm{choice}(\mathbf{x}_{t_{0}}, \mathbf{x}_{t_{1}}, \mathbf{x}_{t_{2}}, \ldots, \mathbf{x}_{t_{i-1}}); \Theta^{\prime}), \label{5}
\end{equation}
where $\mathbf{O}_{SAR}$ represents the prediction order of SAR, in which we can decide each $t_{i}$ freely. And $\mathrm{choice}$ function chooses partial input by masking the rest. If nodes of a directed graph represent elements in the sequence and edges represent dependencies in SAR, then they form a \textbf{dependency} \textbf{graph}. It is easy to prove that the dependency graph is a DAG, and the prediction order is one of its topological sorting arrays. Note that a DAG may have different sorting arrays. They are all valid prediction orders because some nodes share the same dependencies and can be generated simultaneously.
 
Error accumulates through edges of the dependency graph, so we can constraint error accumulation by deliberately designing the $\textbf{O}_{SAR}$ and the $\mathrm{choice}$ function, which decides the graph topology.
% \begin{equation}
%     \mathbf{e}_{t_{i}} = \mathrm{f_{e}}^{\prime}(\mathrm{choice}(\mathbf{e}_{t_{0}}, \mathbf{e}_{t_{1}}, \mathbf{e}_{t_{2}}, \ldots \mathbf{e}_{t_{i-1}})) \label{6}
% \end{equation}
% so we can constraint error accumulation by deliberately designing the $\textbf{O}_{SAR}$ and the $\mathrm{choice}$ function, which decides the topology of the dependency graph. The original dependency graph is a complete DAG, which contains every possible edges, so the error accumulates very fast.

\subsubsection{Pipeline}
We instantiate a dependency graph for motion interpolation, representing a pipeline of 3 stages: keyframe interpolation, frame-by-frame generation, and smoothing.

\noindent \textbf{Keyframe Interpolation}
Frames generated earlier generally have a minor error, making them the best reference for future generations. We select several keyframes, which equally split the entire sequence. Then generating the frames between adjacent keyframes is a smaller-size motion interpolation problem. It is a divide-and-conquer strategy, which can be done recursively by selecting multiple levels of keyframes. For simplicity, we only select one level in the paper.

\noindent \textbf{Frame-by-Frame Generation}
After sub-problems become simple enough, we solve them by a left-to-right autoregression called frame-by-frame generation. This stage ensures the continuity of the sub-problem because of adjacent dependencies.
Keyframe interpolation and frame-by-frame generation can dilute error accumulation in two directions. First, error accumulates through levels of keyframes. Then, it flows through intervals of minimal sub-problems. 

\noindent \textbf{Smoothing}
We apply a global smoothing process by re-generating the sequence depending on all the previous predictions, which is a parallel generation. This stage can also be integrated into the dependency graph by duplicating the previous nodes.

\subsection{Backbone of the Regressor} \label{subsection:model}

As shown in Fig.~\ref{fig:overview}, the regressor consists of (1) Spatial Pose Encoder, (2) Temporal Decoder, (3) Spatial Pose Decoder, and it work under the guidance of the dependency graph that controls the FDAM.

\subsubsection{Spatial Pose Encoder and Motion Embedding}
The model receives the motion sequence $\mathbf{P}_\mathrm{inp} \in \mathbb{R}^{N \times J \times 3}$ as input.
To handle 3D motion sequence, we feed each pose $\mathbf{p}^{(i)}_\mathrm{inp} \in \mathbb{R} ^ {J \times 3} $ into a Spatial Transformer \cite{Vaswani2017} Encoder, which conducts attention between every two joints, receiving an embedding matrix $\mathbf{E}_\mathrm{inp} = [\mathbf{e}_\mathrm{inp}^{(0)}, \mathbf{e}_\mathrm{inp}^{(1)}, \mathbf{e}_\mathrm{inp}^{(2)}, \ldots, \mathbf{e}_\mathrm{inp}^{(N-1)}]^{\mathrm{T}}$, where each embedding $\mathbf{e}_\mathrm{inp}^{(i)}$ is a $J \cdot D$ dimensional vector. Then we add sinusoidal position embedding and feed it to the Temporal Decoder.

\subsubsection{Temporal Decoder}
\label{subsubsec:decoder}
The Temporal Decoder contains several blocks of GPT-2 \cite{Alec2019} with the FDAM Multi-Head Attention. The Temporal Decoder takes $\mathbf{E}_\mathrm{inp}$ as input and outputs a sequence of embedding $\mathbf{E}_\mathrm{gen}$ of the same size.

In the FDAM Multi-Head Attention layer, the FDAM inside it can control which input frames are used or masked for generating each $\mathbf{e}_\mathrm{gen}^{(i)}$, which realizes the $\mathrm{choice}$ function of Equation \eqref{5}.

Following is the algorithm for generating FDAM. For a specific dependency graph, find one of its topological sort arrays $\mathbf{O}_\mathrm{SAR}$. For every element $O_{i}$ in the $\mathbf{O}_\mathrm{SAR}$, find all nodes from which have edges $(x_{j}, O_{i})$ in the graph. Then for each $x_{j}$, set $\mathbf{FDAM}[O_{i-1}, x_{j}] = 1$ (Let $O_{0}=0$). So the FDAM is the transposed adjacent matrix of the dependency graph which is shuffled by a function: $$\mathbf{FDAM}_{O_{i-1}}=\mathbf{AdjacentMatrix}^\mathrm{T}_{O_{i}}.$$

\subsubsection{Spatial Pose Decoder and Shuffled Output} \label{subsubsection:shuffled}
% 先解释一下，为什么output order是乱序的
% 然后定义一个shuffle函数，来描述这个乱序对应的过程
We use an MLP as the Spatial Pose Decoder to decode the embedding $\mathbf{E}_\mathrm{gen}$ to an output pose sequence $\mathbf{P}_\mathrm{gen}$ of the same size as $\mathbf{P}_\mathrm{inp}$.
Because of the SAR, the output is in shuffled order as described in subsubsection~\ref{subsubsec:decoder}. For every generation, we receive one valid element $\mathbf{p}^{(\mathbf{O}_{i})}$ from the active path $\mathbf{p}_\mathrm{gen}^{(\mathbf{O}_{i-1})}$. Then we put it back and run for another iteration.

\subsection{Training and Inference}
The training method contains two steps. In the first step, we apply a teacher-forcing strategy that facilitates convergence. We optimize MSE loss between the result and ground truth. In the second step, the model's output is entirely generated from the start and end frame, improving test-time performance. We block the gradients of the model and use the trained model to operate the first two stages. Then we feed the trajectory to the network, generating the whole sequence in parallel for the second time. Then We optimize MSE loss between the generated frames and the ground truth. The inference method works in the same way as the second training step.
\section{Expriment}
\label{sec:pagestyle}

\subsection{Dataset}
We adopt SMPL+H \cite{Loper2015, Romero2017} skeleton and conduct experiments on a dataset constructed from the AMASS dataset \cite{Mahmood2019}. The raw data contains various actions and motion sequences at different frame rates. We follow previous work \cite{harvey2020} to cut a long motion sequence into small pieces by sliding a window. We set the sequence length to 31. However, to enforce our model's generality, we embrace motions of various amplitudes and design different sliding windows for different framerates. For a high framerate, we set the window lengths assigned to both 31 and 62 frames, then downsample the latter one. Totally, we have a 166,696 sliced sequence. The constructed dataset is split into training, validation, and test splits consisting of roughly 70\%, 10\%, and 20\% of the samples, respectively.

\noindent\textbf{Implementation Details}
For pose embedding, we use a Transformer encoder which consists of 4 blocks with embedding dimensions of size 1248 (24 for each joint) and 12 heads attention layers. 
For Temporal Decoder, we use 6 blocks with 8 heads of attention.
For prediction order, we select keyframes set [1, 9, 19, 29].
% At teacher-forcing training, we train our network with Adam optimizer at a learning rate of 6e-4 and weight decay of 0.01. At the smoothness training, we train our network with Adam optimizer at a learning rate of 1e-5 and weight decay of 0.01. All models are trained and tested with PyTorch and the Nvidia RTX 2080Ti GPU.

\subsection{Evaluate Metrics}
\textbf{Reconstruction Loss}
We leverage the standard metrics Mean Per Joint Angle Error (MPJAE) and Mean Per Joint Position Error (MPJPE) to evaluate motion interpolation. 

\noindent\textbf{Neighbour L2 Distance \cite{Wang2021}}
An indicator of both naturalness and smoothness, which can be viewed as the speed of motion. The closer the measured value is to the ground truth value, the better.

\noindent\textbf{NPSS}
We adopt  Normalized Power Spectrum Similarity (NPSS) \cite{gopalakrishnan2019}, which is proved to be highly correlated to the human assessment of motion quality.

\begin{table}[t]
	\centering
		\caption{Quantitative results against several baseline methods and ablation study results. For all metrics, the lower, the better. The Neighbor L2 Distance is shown as the value minus the ground truth value (full model value) for quatitative (ablation) results.}
	\setlength\tabcolsep{7pt}
	\begin{tabular}{l c c c c c}\hline
		Model&MPJAE&MPJPE&Neigh Dist&NPSS\\\hline
		QN \cite{Pavllo2019}
		&0.3922 &0.6759 &105.9001      &2.868\\
		BERT \cite{duan2022}
		&0.0153 &0.0466 &-1.9475       &0.095\\
		CAE \cite{kaufmann2020}
		&0.0247 &0.0852 &-1.9628      &0.121\\
		SLERP \cite{shoemake1985}
		&0.0103 &0.0406 &-5.2662      &0.093\\
        
		Ours
		& \textbf{0.0102} &\textbf{0.0365} &\textbf{-1.0368}       &\textbf{0.089}\\
		\hline
        Full Model
		& 0.0102 &0.0365 &0   &0.089\\
		Original AR&0.1847&0.3461&89.395&6.594\\
		Binary Search&0.0152 &0.0584 &1.3489&0.107\\
		w/o smoothing&0.0121 &0.0387 &2.0001&0.096\\
		\hline
	\end{tabular}
	\label{tab:results}
\end{table}

\subsection{Results}

We compare our model with four methods, including neural networks and mathematical functions. For neural networks, We choose Quaternet (QN) proposed by \cite{Pavllo2019}, a unified BERT-like model (BERT) proposed by \cite{duan2022} and a convolutional autoencoder (CAE) proposed by \cite{kaufmann2020} as baselines. The selection covers the main types of motion modeling solutions and contains each type's state-of-the-art methods.
Additionally, we use spherical linear interpolation (SLERP) as our baseline of mathematical methods.

\noindent \textbf{Quantitative Results}
The performance of different methods tested on the test dataset is summarized in Table~\ref{tab:results}. 

Quaternet\cite{Pavllo2019}, transferred from the motion prediction task, fails to work since it is autoregressive.
Our model outperforms other deep learning models by a large margin on reconstruction losses and NPSS. And our model is the closest to the ground truth value when evaluating the Neighbor L2 Distance. The other models have a lower value than the ground truth value, which indicates that their results are more monotonous.
SLERP predicts an average trajectory of the motion space, resulting in advantages in metrics values.  Therefore, its good metric values can not reflect its visual quality.
% In terms of objective metrics, SLERP is the second best one, but the non-deep learning model will predict the average trajectory of the motion space, resulting in advantages in metrics values. Because of their irrelevance to human motion background, the non-deep learning methods generate unnatural motions seen in the qualitative results.

\begin{figure}[t]
\centering
    \includegraphics[scale=0.5]{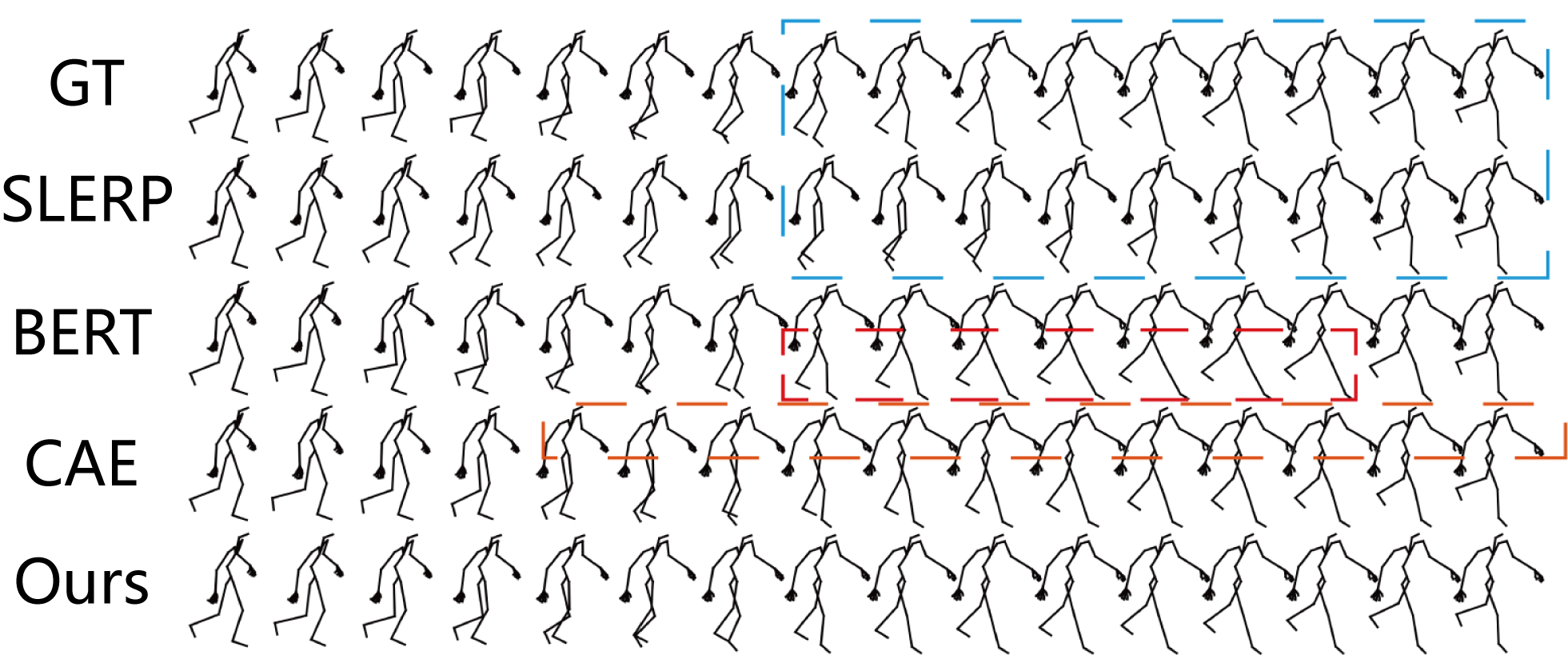}
    \caption{Qualitative Comparison. The expected motion is taking a step forward. GT stands for the ground truth. Dotted rectangles in the figure mark the defects of methods.}
    \label{fig:baseline}
\end{figure}

\noindent \textbf{Qualitative Results}
We randomly select 20 samples from the test dataset. We demonstrate the most representative sample where the motion is taking a step forward in Fig.~\ref{fig:baseline}.
 
The SLERP interpolation is weaker in naturalness than other models. And the uniform rotation of the knees in SLERP's result leads to penetration with the ground.
The autoencoder model \cite{kaufmann2020} generates a trembling upper body because it is a local-oriented model based on CNN. The jitter can be seen more clearly in the video.
Our model and the BERT-like model \cite{duan2022} have better results in motion interpolation since both can generate coherent actions. The over-smoothed problem of the BERT-like model from its linear interpolation input has not been fully solved. It can be seen in the figure that the actor of the BERT-like model seems to be throwing his legs powerlessly. Compared with the Bert-like model, our results show the strength of the human body and are closer to the ground truth.

\subsection{Ablation Study}
\label{subsec:ablation}
We conduct an ablation study on AMASS dataset to test each pipeline function, presented in Table~\ref{tab:results}. The results show that each component of our model contributes to the performance.

Firstly we train the model with the original autoregression to evaluate the importance of the SAR. The model fails to work as expected. As the frame-by-frame generation progresses, the motion speed gradually decreases due to error accumulation. Finally, the model tends to predict no movement visually and fails to connect the end pose. Secondly, the SAR with a dependency graph that does not contain the frame-by-frame generation is tested. We follow the order of binary search, which recursively generates the middle frame of an interval. The output becomes incoherent because of the scarcity of local dependencies. Thirdly, we train the model without the smoothing step. The growth of Neighbour L2 Distance value shows that the smoothing step significantly improves smoothness.

\section{Conclusions}
% In this work, we propose a novel approach, shuffled autoregression, which assembles a new pipeline for motion interpolation into an end-to-end framework. The approach can be extended to interpolation problems in other fields. We further implement the approach to GPT architecture by proposing a module called Flexible Dependency Attention Mask (FDAM). And we show that the dependency relations of such architecture is actually a directed acyclic graph (DAG) and the prediction order is one of its topological sorting arrays. We manipulate FDAM to change topology which alleviates error accumulation. Experiments on the AMASS dataset show that our model outperforms other methods for similar tasks.
In this work, we propose the novel Shuffled AutoRegression (SAR) for motion interpolation. Based on the observation that the dependency graph of SAR is a directed acyclic graph, we further propose an idea for constructing a specific kind of dependency graph. The topology of the graph represents a pipeline containing three stages: (1) keyframes interpolation, (2) frame-by-frame generation, and (3) smoothing. Additionally, we devise the Flexible Dependency Attention Mask (FDAM) and plug this module into our backbone regressor. Our framework alleviates the error accumulation problem and generates consecutive and natural motions. Experiments on the AMASS dataset show that our model outperforms other methods for similar tasks. Our framework can also be extended to multiple keyframes' interpolation tasks and interpolation in other fields.
% To evaluate the importance of SAR, we train the model with the ordinary mask (in sequential order). As the frame-by-frame generation progresses, the motion speed gradually decreases due to error accumulation. Finally, the model shows a tendency to predict no movement visually. As shown in Figure~\ref{fig:ablation}, the arm of the actor moves a little bit at first and then remains unmoved.

% Then We analyze the influence of the keyframes number $N_k$. For different values of $N_k$, we equally split the range of 31 frames. It's a trade-off that the more keyframes, the less accurate the generation of the keyframes, while the fewer keyframes, the less accurate the generation inside subintervals.

% We design a prediction order which does not contain the frame-by-frame generation. We follow the order of binary search, which recursively generates the middle frame of an interval. As shown in Figure~\ref{fig:ablation}, the output is incoherent, with fast movements connecting to slow ones. It can also be seen on the Neighbor L2 Distance in Table~\ref{tab:ablation}.

% We also train the model without the smoothing step. As shown in Table~\ref{tab:ablation}, the smoothing step significantly lowers the Neighbor L2 Distance, which improves smoothness. In Figure~\ref{fig:ablation}, the actor's arms are trembling, which significantly damages the visual quality.

% Finally, we test the contribution of the Spatial Pose encoder, where we leverage a Transformer \cite{Vaswani2017} encoder. We replace it with an MLP of one layer, resulting in bad performance shown in Table~\ref{tab:ablation}.
\vspace{8pt}
\noindent \textbf{Acknowledgments:}
This work is supported by the National Key R\&D Program of China under Grant No.2021QY1500, the state key program of the National Natural Science Foundation of China (NSFC) (No.61831022).

\vfill\pagebreak

\bibliographystyle{IEEEbib}
\bibliography{template}

\end{document}